\documentclass[letterpaper]{article} % DO NOT CHANGE THIS
\usepackage{aaai2027}
\usepackage[hyphens]{url}  % DO NOT CHANGE THIS
\usepackage{graphicx} % DO NOT CHANGE THIS
\usepackage{natbib}  % DO NOT CHANGE THIS AND DO NOT ADD ANY OPTIONS TO IT
\usepackage{caption} % DO NOT CHANGE THIS AND DO NOT ADD ANY OPTIONS TO IT
\usepackage{subcaption}
\usepackage{amsmath}
\usepackage{amssymb}
\usepackage{booktabs}
\usepackage{array}
\usepackage{algorithm}
\usepackage{algorithmic}

\newcommand{\vx}{\mathbf{x}}
\newcommand{\vr}{\mathbf{r}}
\newcommand{\vc}{\mathbf{c}}
\newcommand{\vmu}{\boldsymbol{\mu}}
\newcommand{\vu}{\mathbf{u}}
\newcommand{\vv}{\mathbf{v}}
\newcommand{\vzero}{\mathbf{0}}

\newcommand{\norm}[1]{\left\lVert #1 \right\rVert_2}
\newcommand{\normalize}[1]{\operatorname{norm}\!\left(#1\right)}
\newcommand{\topk}{\operatorname{TopK}}
\newtheorem{proposition}{Proposition}

\title{PRQ-KMeans: Projection Residual Quantization for Semantic ID Tokenization}
\author{Anonymous Submission}
\affiliations{}
\author{
Yunxiao Luo\thanks{Equal contribution.},
Siyuan Wang\thanks{Equal contribution.},
Ben Chen\thanks{Corresponding author.},
Chenyi Lei,
Qingpeng Cai
}

\affiliations{
\textsuperscript{1}Kuaishou Technology\\
}

\begin{document}

\maketitle

\begin{abstract}
Semantic identifiers (SIDs) represent entities as hierarchical token sequences for generative retrieval and recommendation.
Residual-quantization tokenizers construct these sequences by selecting a codeword at each level and passing a residual to the next.
We view this process as \textbf{progressive commonality removal}: each token captures a component shared within its group, while later tokens should model the remaining differences.
This view reveals three limitations: a corpus-wide shared component can consume first-level capacity, hard assignment ignores graded similarities to nearby codewords, and full-codeword subtraction can leave variation along the selected-codeword direction in the next residual.
We therefore develop our solution in the post-hoc setting, where residual construction is not constrained by input reconstruction.
Specifically, we propose \textbf{PRQ-KMeans}, which removes the global-mean component, refines centroids with Top-$k$ similarity-weighted updates, and replaces full-codeword subtraction with a projection residual that removes each representation's selected-centroid component.
Experiments on a large-scale industrial search dataset and four public recommendation benchmarks show that PRQ-KMeans achieves the strongest overall performance among the evaluated tokenizers, including gains of up to 7.4\% in HitRate and 11.8\% in MRR on the industrial dataset.
\end{abstract}

\section{Introduction}

Semantic identifiers (SIDs) replace atomic entity identifiers with short token sequences for generative retrieval \cite{tay2022dsi,wang2022nci,sun2023genret} and recommendation \cite{rajput2023tiger,singh2023semanticids}.
Hierarchical SIDs organize these tokens from coarse to fine: related entities can share early tokens, while later tokens distinguish entities that remain under the same prefix.
Constructing such an identifier involves two coupled decisions at each level: which token represents the entity at the current position, and what representation is passed forward to construct the next token.

Residual quantization provides a common framework for making these decisions.
At each level, it selects a codeword for the current token, subtracts that codeword from the current representation, and passes the resulting residual to the next level.
Two major tokenizer paradigms use this recursion.
VAE-based methods \cite{lee2022rqvae,rajput2023tiger,wan2026r3vae} learn hierarchical SIDs through a residual-quantization autoencoder.
Post-hoc methods (e.g., RQ-KMeans and RQ-GMM) \cite{luo2024qarm,deng2025onerec,tong2026rqgmm} instead fit hierarchical codebooks directly over existing embeddings.
In both paradigms, the residual becomes the representation processed at the next level.
We refer to the component that remains along the selected-codeword direction after subtraction as residual carryover.

\begin{figure}[t]
  \centering
  \includegraphics[width=\columnwidth]{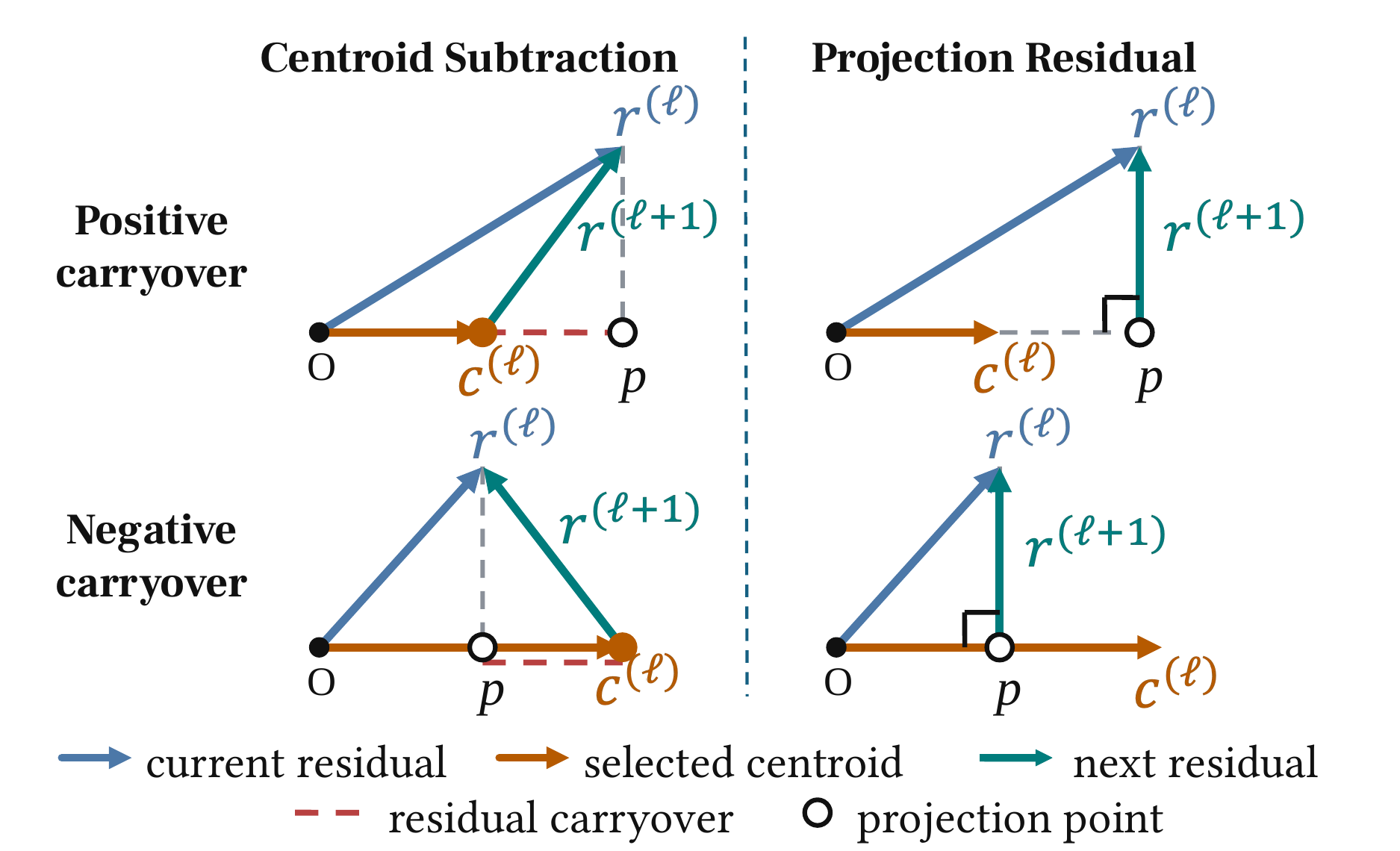}
  \caption{Full-centroid subtraction and projection residuals.
  Full-centroid subtraction may leave positive or negative residual carryover along the selected-centroid direction, whereas projection removes this component and yields a residual orthogonal to that centroid.}
  \label{fig:residual-orthogonality}
\end{figure}

Since each later token uses the preceding residual, residual construction determines what information passes across levels.
We therefore view residual quantization as \textbf{progressive commonality removal}: the selected codeword represents a component shared within the current group, while the residual should retain the differences that remain for later tokens.
This view reveals three challenges.
\textbf{First}, commonality can exist before any cluster-level token is formed.
A component shared across the embedding collection may dominate the initial representations.
Because it is broadly shared, this component contributes little to distinguishing entities, yet the first codebook may spend part of its capacity modeling this global background rather than coarse differences among entities.
\textbf{Second}, hard assignment limits how accurately a codeword can summarize its group.
Each representation updates only its selected codeword, even when it is also similar to nearby candidates, discarding graded similarity around assignment boundaries.
\textbf{Third}, full-codeword subtraction does not generally remove the selected-codeword component accurately from every representation.
As Figure~\ref{fig:residual-orthogonality} illustrates, representations assigned to the same codeword may contain different amounts of this component, although the same codeword is subtracted from all of them.
This mismatch can leave a positive or negative component along the selected direction, causing the next codebook to revisit variation already represented by the current token.
Our preliminary study finds this component measurable in industrial RQ-KMeans residuals and shows that retaining more of it monotonically reduces next-level SID-prefix utilization (Section~\ref{sec:preliminaries}).

VAE-based tokenizers can reshape their latent representations and codebooks through reconstruction and auxiliary SID objectives \cite{wang2024letter,zhu2024cost,wan2026r3vae}.
Their residuals, however, are constrained by additive reconstruction: everything left after subtracting the current codeword is passed to later codewords, even a component along the same direction.
Discarding that component independently would alter the codeword sum used for reconstruction.
Post-hoc tokenizers instead operate on existing embeddings without requiring the codewords to reconstruct the input, allowing the residual to preserve only the differences needed by later SID levels.
We develop our solution in this post-hoc setting using RQ-KMeans.

To address these challenges, we propose \textbf{PRQ-KMeans}, a post-hoc hierarchical SID tokenizer built around projection residuals.
To keep the corpus-wide component from consuming first-level capacity, PRQ-KMeans removes each input representation's projection onto the global mean before fitting the first codebook.
To better estimate the component shared within each cluster, Top-$k$ soft refinement lets every representation contribute to several nearby centroids in proportion to their cosine similarities, rather than updating only its most similar centroid.
To prevent the selected-centroid component from being passed to the next level, PRQ-KMeans removes each representation's own projection onto its selected centroid after hard assignment, rather than subtracting the full centroid vector.
The resulting residual is orthogonal to that centroid and is passed forward to fit the next codebook.
Together, these operations implement progressive commonality removal from global to local.

Our main contributions are threefold.
(1) We introduce the view of hierarchical SID construction as progressive commonality removal and identify three challenges in determining what is passed between successive token levels.
(2) We propose PRQ-KMeans, which combines global component removal and Top-$k$ soft centroid refinement with a projection residual that removes this component separately for each representation.
(3) We demonstrate the effectiveness of PRQ-KMeans through extensive evaluation on industrial search and four public recommendation benchmarks.

\section{Related Work}

\subsection{Semantic IDs in Generative Retrieval and Recommendation}

Semantic IDs represent entities as short sequences of discrete tokens. For generative retrieval, DSI \cite{tay2022dsi} establishes the paradigm of directly generating document identifiers, and NCI \cite{wang2022nci} extends this paradigm with hierarchical semantic identifiers. GenRet \cite{sun2023genret}, GLEN \cite{lee2023glen}, and MERGE \cite{zhang2025merge} incorporate semantic, lexical, or relevance signals into identifier construction. OneSearch \cite{chen2025onesearch} applies hierarchical SIDs to large-scale generative e-commerce search. For generative recommendation and ranking, TIGER \cite{rajput2023tiger}, Semantic IDs for Ranking \cite{singh2023semanticids}, and OneRec \cite{deng2025onerec} use hierarchical SIDs to organize the output space. LETTER \cite{wang2024letter}, CoST \cite{zhu2024cost}, ETEGRec \cite{liu2025etegrec}, DIGER \cite{fu2026diger}, and UniSID \cite{jiang2026unisid} further incorporate collaborative, contrastive, differentiable, or end-to-end supervision into SID learning. For embedding-derived SIDs, the tokenizer determines how continuous entity representations are mapped to discrete token sequences.

\subsection{Quantization Methods for Hierarchical SIDs}

Residual quantization constructs a hierarchical SID by successively quantizing the residual passed from one level to the next. RQ-VAE \cite{lee2022rqvae,rajput2023tiger} and R3-VAE \cite{wan2026r3vae} use autoencoders and recursive residual quantization to learn hierarchical SIDs. Post-hoc methods fit the hierarchy over existing embeddings: QARM \cite{luo2024qarm} and OneRec \cite{deng2025onerec} use RQ-KMeans, while RQ-GMM \cite{tong2026rqgmm} uses probabilistic mixture components. Beyond residual quantization, parallel methods generate multiple codes from separate vector subspaces or token branches, including PQ \cite{jegou2011pq}, OPQ \cite{ge2013opq}, LLaDA-Rec \cite{shi2025lladarec}, and ACERec \cite{xia2026acerec}. Moreover, hybrid tokenizers, including OneSearch \cite{chen2025onesearch}, QARM V2 \cite{xia2026qarmv2}, and Quantizing Intent \cite{choi2026quantizing}, combine residual and non-residual quantizers to capture hierarchical commonality and finer distinctions. Recently, methods such as DOS \cite{yin2026dos}, ReSID \cite{liang2026resid}, and HHQ \cite{wang2026hhq} reshape the quantization space through orthogonal or hyperspherical structures. However, these methods leave the cross-level residual unexplored. Under full-centroid subtraction, the residual can still carry a selected-centroid component into the next codebook, making later levels revisit variation represented earlier. PRQ-KMeans addresses this gap by modeling this carryover and removing it through per-representation projection.

\section{Preliminaries and Motivation}
\label{sec:preliminaries}

\subsection{RQ-KMeans for Hierarchical SIDs}

In RQ-KMeans, each codeword is a K-Means centroid.
Let $\mathcal{X}=\{\vx_i\}_{i=1}^{N}\subset\mathbb{R}^{d}$ be the embeddings used to fit the tokenizer. A depth-$L$ RQ-KMeans tokenizer learns one codebook $\mathcal{C}^{(\ell)}=\{\vc_j^{(\ell)}\}_{j=1}^{K_\ell}$ at each level and maps entity $i$ to the SID $(z_i^{(1)},\ldots,z_i^{(L)})$. Starting from $\vr_i^{(1)}=\vx_i$, Euclidean K-Means alternates between nearest-centroid assignment and centroid update:
\begin{equation}
\begin{aligned}
z_i^{(\ell)}
&=\arg\min_j \norm{\vr_i^{(\ell)}-\vc_j^{(\ell)}}^2,\\
\vc_j^{(\ell)}
&\leftarrow
\frac{1}{|\mathcal{I}_j^{(\ell)}|}
\sum_{i\in\mathcal{I}_j^{(\ell)}}\vr_i^{(\ell)}.
\end{aligned}
\label{eq:rq-kmeans-fitting}
\end{equation}
where $\mathcal{I}_j^{(\ell)}=\{i:z_i^{(\ell)}=j\}$. After fitting, $z_i^{(\ell)}$ becomes the token at level $\ell$, and RQ-KMeans constructs the next-level representation by subtracting the selected centroid:
\begin{equation}
\vr_i^{(\ell+1)}
=\vr_i^{(\ell)}-\vc_{z_i^{(\ell)}}^{(\ell)}.
\label{eq:standard-rq}
\end{equation}
Some implementations additionally apply the in-place update $\vr_i^{(\ell+1)}\leftarrow\normalize{\vr_i^{(\ell+1)}}$ before the next level. Following previous work \cite{chen2025onesearch,chen2026onesearchv2}, the industrial analysis below uses this normalized variant. Here, $\normalize{\vv}=\vv/\norm{\vv}$ for nonzero $\vv$. In reconstruction-oriented quantization, the residual is the approximation error left after selecting a centroid; in hierarchical SID tokenization, it also determines the representation available to the next level. Under \emph{progressive commonality removal}, the selected centroid summarizes information shared within its cluster, while the residual should preserve the remaining differences. We next examine whether full-centroid subtraction achieves this goal for each representation.

\subsection{Residual Carryover after Centroid Subtraction}

A centroid is shared by all representations assigned to its cluster, although these representations may contain different amounts of its directional component. Consider one representation--centroid pair and abbreviate it as $\vr$ and $\vc\neq\vzero$. Write $\vc=\rho\hat{\vc}$, where $\hat{\vc}=\vc/\norm{\vc}$ and $\rho=\norm{\vc}$. Let $\alpha=\vr^\top\hat{\vc}$ and $\vu=\vr-\alpha\hat{\vc}$, so that $\vu\perp\hat{\vc}$. Then
\begin{equation}
\begin{aligned}
\vr &= \alpha\hat{\vc}+\vu, \\
\vr-\vc &= (\alpha-\rho)\hat{\vc}+\vu.
\end{aligned}
\label{eq:centroid-decomp}
\end{equation}
Full-centroid subtraction applies the shared coefficient $\rho$, whereas representation $\vr$ contains the instance-specific coefficient $\alpha$. When $\alpha\neq\rho$, the residual retains $(\alpha-\rho)\hat{\vc}$ along the selected centroid direction; we call this component \emph{residual carryover}. Figure~\ref{fig:residual-orthogonality} illustrates the positive- and negative-carryover cases. An arithmetic-mean centroid matches the average coefficient within its assigned cluster, but not necessarily the coefficient of every representation. Appendix~\ref{app:proofs} provides the corresponding identity and derivation.

\subsection{Empirical Motivation}

\begin{figure}[t]
  \centering
  \begin{subfigure}[t]{0.485\columnwidth}
    \centering
    \includegraphics[width=\linewidth]{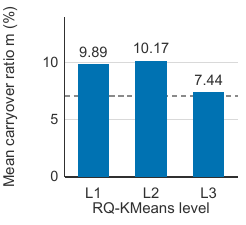}
    \caption{Residual carryover}
  \end{subfigure}\hfill
  \begin{subfigure}[t]{0.485\columnwidth}
    \centering
    \includegraphics[width=\linewidth]{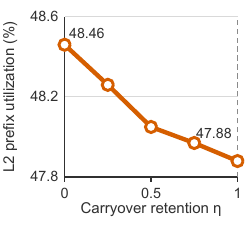}
    \caption{Controlled L2 utilization}
  \end{subfigure}
  \caption{Empirical motivation on the industrial dataset. (a) Mean residual carryover ratio at each RQ-KMeans level. The dashed line marks the expected absolute cosine between two independent isotropic directions in 128 dimensions. (b) L2 prefix utilization after retaining a fraction $\eta$ of the L1 carryover. L1 assignments stay fixed; L2 is refit for every $\eta$.}
  \label{fig:carryover-retention-utilization}
\end{figure}

We first measure residual carryover using a three-level RQ-KMeans tokenizer fitted to representations from the industrial dataset, with codebook sizes $1024$, $512$, and $128$. Section~\ref{sec:experimental-setup} and Appendix~\ref{app:experimental-config} provide the setup. For representation $i$, define the relative carryover magnitude at level $\ell$ as
\begin{equation}
m_i^{(\ell)}
=
\frac{
\left|\alpha_i^{(\ell)}-\rho_i^{(\ell)}\right|
}{
\norm{
\vr_i^{(\ell)}
-
\vc_{z_i^{(\ell)}}^{(\ell)}
}
}.
\label{eq:kuaishou-parallel-fraction}
\end{equation}
Figure~\ref{fig:carryover-retention-utilization}(a) reports mean ratios of 9.89\%, 10.17\%, and 7.44\% at L1, L2, and L3, respectively. For scale, two independent isotropic directions in 128 dimensions have an expected absolute cosine of 7.07\%; Appendix~\ref{app:residual-geometry} provides the derivation. The L1 and L2 means, corresponding to residuals passed to subsequent codebooks, exceed this reference by 2.82 and 3.10 percentage points, respectively, whereas the terminal L3 value is close to it.

We next test how this component affects the immediately following SID level. Keeping the L1 assignments fixed, we retain a fraction $\eta\in[0,1]$ of the L1 carryover:
\begin{equation}
\vr_i^{(2)}(\eta)
=
\normalize{
\eta\left(\alpha_i^{(1)}-\rho_i^{(1)}\right)
\hat{\vc}_i^{(1)}+\vu_i^{(1)}
}.
\label{eq:carryover-retention}
\end{equation}
Here, $\hat{\vc}_i^{(1)}$ is the direction of the selected L1 centroid. The endpoints $\eta=1$ and $\eta=0$ retain the full RQ-KMeans carryover and remove it completely, respectively. For each $\eta$, we refit L2 with the same Euclidean K-Means configuration and measure the fraction of the $K_1K_2$ possible two-token item prefixes assigned to at least one item. Appendix~\ref{app:residual-geometry} provides the equivalent operational update and implementation details.

\begin{figure*}[!t]
  \centering
  \includegraphics[width=\textwidth]{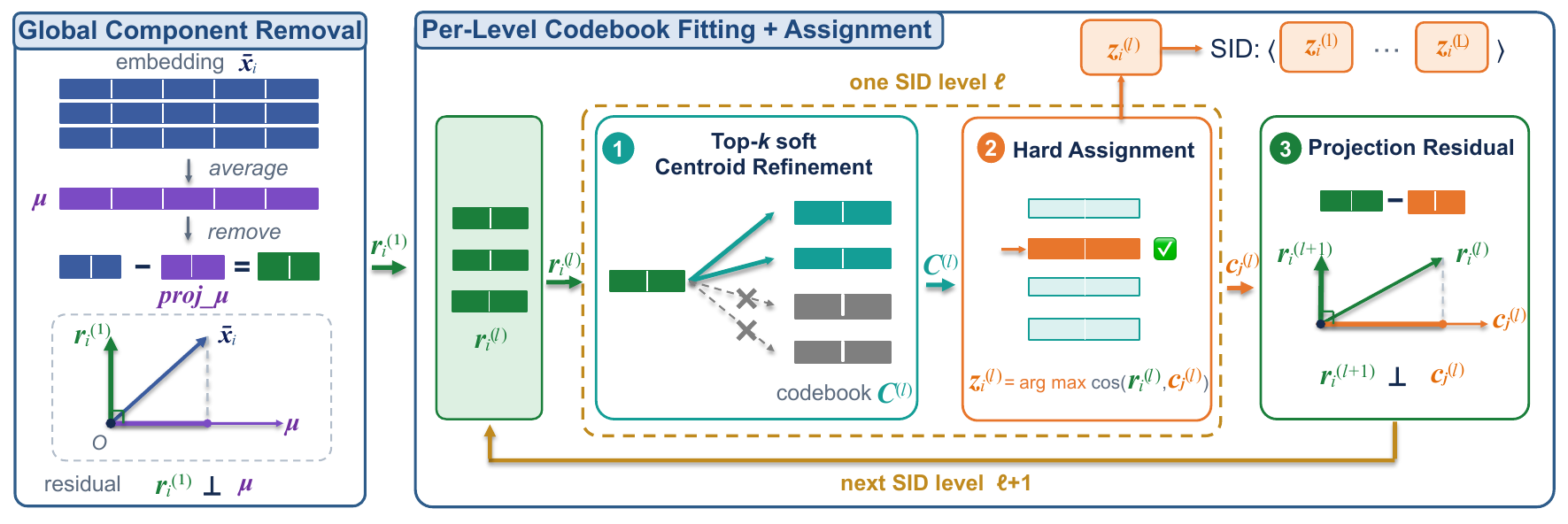}
  \caption{Overview of PRQ-KMeans. PRQ-KMeans initializes the first-level residual by removing the global-mean component. At level $\ell$, Top-$k$ soft centroid refinement fits the codebook from the current residuals. Hard cosine assignment then selects a centroid and emits token $z_i^{(\ell)}$; projection removes the selected-centroid component before the residual proceeds to level $\ell+1$. SID encoding repeats the same hard-assignment and projection recursion using the learned codebooks.}
  \label{fig:dark-overview}
\end{figure*}

Figure~\ref{fig:carryover-retention-utilization}(b) shows that L2 prefix utilization increases monotonically from $47.88\%$ at $\eta=1$ to $48.46\%$ at $\eta=0$.
Under this controlled protocol, reducing selected-centroid carryover allows the immediately following level to use a broader set of prefixes.
This observation motivates constructing the next-level representation by removing the selected-centroid component through projection.

\section{Methodology}
\label{sec:dark}

\subsection{Overview}

Figure~\ref{fig:dark-overview} gives an overview of PRQ-KMeans.
It first removes the global component from the input embeddings.
It then learns the codebooks sequentially: at each level, Top-$k$ soft weights refine the current centroids, hard cosine assignment emits a token, and the projection residual becomes the input used to fit the next codebook.
Once the hierarchy is trained, SID encoding repeats the hard assignment and residual update using the learned global mean and codebooks during inference.
Projection defines the cross-level update, while global removal and soft refinement prepare its input representations and centroids.
Appendix~\ref{app:algorithm} gives the complete algorithm.

\subsection{Projection Residual Construction}

PRQ-KMeans removes the selected-centroid component by requiring the next residual $\vv$ to be orthogonal to the selected centroid $\vc$.
Orthogonality alone, however, does not determine a useful residual; even the zero vector satisfies this constraint.
We therefore choose the orthogonal residual that changes the current representation $\vr$ as little as possible:
\begin{equation}
\vv^\star
=\underset{\vv}{\arg\min}\;\norm{\vv-\vr}^{2}
\quad\text{s.t.}\quad
\vv^\top\vc=0.
\label{eq:orthogonal-residual-opt}
\end{equation}
Solving this problem gives
\begin{equation}
\vv^\star
=\vr-\frac{\vr^\top\vc}{\norm{\vc}^{2}}\vc.
\label{eq:general-projection-residual}
\end{equation}

This solution directly removes the carryover identified in Eq.~\eqref{eq:centroid-decomp}.
Recall that $\vc=\rho\hat{\vc}$, $\alpha=\vr^\top\hat{\vc}$, and $\vr=\alpha\hat{\vc}+\vu$.
The component subtracted in Eq.~\eqref{eq:general-projection-residual} is
\[
\frac{\vr^\top\vc}{\norm{\vc}^{2}}\vc
=
\alpha\hat{\vc}.
\]
Consequently, $\vv^\star=\vr-\alpha\hat{\vc}=\vu$.
Unlike the RQ-KMeans residual $(\alpha-\rho)\hat{\vc}+\vu$, the projection residual contains no remaining component along the selected centroid.

Proposition~\ref{prop:projection-properties} formalizes these guarantees.
\begin{proposition}[Minimum-change orthogonal residual]
\label{prop:projection-properties}
For any $\vr\in\mathbb{R}^{d}$ and nonzero $\vc$, the projection residual $\vv^\star$ in Eq.~\eqref{eq:general-projection-residual} is the unique solution to Eq.~\eqref{eq:orthogonal-residual-opt}.
It satisfies $\vc^\top\vv^\star=0$ and has the smallest Euclidean distance to $\vr$ among all vectors orthogonal to $\vc$.
\end{proposition}

Centroid magnitude does not affect the component removed by Eq.~\eqref{eq:general-projection-residual}.
PRQ-KMeans therefore compares representations and centroids using cosine similarity, which is likewise unaffected by centroid magnitude.
The same cosine scores are used for Top-$k$ soft refinement and hard assignment.
The complete proofs are provided in Appendix~\ref{app:proofs}.

\subsection{Global Component Removal}

A component shared across the entire embedding collection can dominate the first partition.
Standard K-Means uses the mean of each cluster as its centroid, summarizing information shared by the representations in that cluster.
PRQ-KMeans applies the same idea before the first SID level: it uses the mean of all input representations to summarize information shared across the collection.
We write $\bar{\vx}_i=\normalize{\vx_i}$ for the L2-normalized input embedding.
Their global mean is
\begin{equation}
\vmu=\frac{1}{N}\sum_{i=1}^{N}\bar{\vx}_i.
\label{eq:global-mean}
\end{equation}
Before emitting any token, PRQ-KMeans removes the component along this mean and obtains the first-level representation:
\begin{equation}
\vr_i^{(1)}
=
\normalize{
\bar{\vx}_i-
\frac{\bar{\vx}_i^\top\vmu}{\norm{\vmu}^{2}}\vmu
}.
\label{eq:global-removal}
\end{equation}
The global mean is learned once during tokenizer fitting and reused during encoding.
Equation~\eqref{eq:global-removal} produces the representations used to fit the first codebook; subsequent levels apply the same residual construction to the selected centroid.

\subsection{Top-$k$ Soft Centroid Refinement}

Because the residual update depends on the selected centroid, reliable centroid estimation matters.
Hard K-Means uses each representation to update only its assigned centroid, ignoring its similarity to nearby centroids around a cluster boundary.
During each fitting iteration, PRQ-KMeans computes the cosine scores and selects the Top-$k$ candidate centroids:
\begin{equation}
\begin{aligned}
s_{ij}^{(\ell)}
&=
\operatorname{cos}
\left(\vr_i^{(\ell)},\vc_j^{(\ell)}\right),\\
\mathcal{N}_k(\vr_i^{(\ell)})
&=
\topk_{j\in\{1,\ldots,K_\ell\}}s_{ij}^{(\ell)}.
\end{aligned}
\end{equation}
Within this candidate set, the soft weight is
\begin{equation}
w_{ij}^{(\ell)}=
\frac{\exp\!\left(\beta s_{ij}^{(\ell)}\right)}
{\sum_{t\in\mathcal{N}_k(\vr_i^{(\ell)})}
\exp\!\left(\beta s_{it}^{(\ell)}\right)}
\end{equation}
for $j\in\mathcal{N}_k(\vr_i^{(\ell)})$, and zero otherwise, where $\beta>0$ is the softmax concentration (inverse-temperature) parameter; larger $\beta$ produces sharper weights.

Each centroid is updated using the representations for which it is a candidate:
\begin{equation}
\vc_j^{(\ell)}\leftarrow
\frac{\sum_i w_{ij}^{(\ell)}\vr_i^{(\ell)}}
{\sum_i w_{ij}^{(\ell)}}.
\label{eq:centroid-update}
\end{equation}
Restricting the weights to the $k$ nearest centroids keeps each update local while allowing a boundary representation to contribute to nearby candidates.
The score, weight, and centroid-update steps are repeated for a fixed number of fitting iterations.
Soft weights are used only while refining the current codebook.
After refinement, hard assignment emits the token and constructs the residual passed to the next level.

\subsection{Sequential Codebook Fitting and SID Encoding}

In the recursion, Eq.~\eqref{eq:general-projection-residual} is applied with $\vr=\vr_i^{(\ell)}$ and $\vc=\vc_{z_i^{(\ell)}}^{(\ell)}$.
After fitting $\mathcal{C}^{(\ell)}$, PRQ-KMeans assigns each fitting representation to one centroid using hard cosine assignment:
\begin{equation}
z_i^{(\ell)}=
\arg\max_j
\operatorname{cos}
\left(\vr_i^{(\ell)},\vc_j^{(\ell)}\right).
\label{eq:hard-assignment}
\end{equation}
The selected index $z_i^{(\ell)}$ gives the current token.
Writing the selected centroid as $\vc_i^{(\ell)}=\vc_{z_i^{(\ell)}}^{(\ell)}$, PRQ-KMeans removes its component and normalizes the result in one update:
\begin{equation}
\vr_i^{(\ell+1)}
=
\normalize{
\vr_i^{(\ell)}
-
\frac{
\vr_i^{(\ell)\top}
\vc_i^{(\ell)}
}{
\norm{\vc_i^{(\ell)}}^{2}
}
\vc_i^{(\ell)}
}.
\label{eq:local-projection-recursion}
\end{equation}
For a nonzero projection residual, the normalization in Eq.~\eqref{eq:local-projection-recursion} is well defined; vector $\vr_i^{(\ell+1)}$ is used to fit $\mathcal{C}^{(\ell+1)}$.
Repeating this process learns the hierarchy level by level.

Once fitting is complete, SID encoding reuses the learned global mean and codebooks, repeating the same hard assignment, token emission, projection, and normalization steps without further updates.

\section{Experiments}
\label{sec:experiments}

\subsection{Experimental Setup}
\label{sec:experimental-setup}

\paragraph{Datasets.}
We primarily evaluate PRQ-KMeans on a publicly released industrial e-commerce search dataset from prior work~\cite{li2026kuaisearch}. It contains approximately 7.8 million items and 9.0 million training-query records for tokenizer fitting.
Following OneSearch \cite{chen2025onesearch}, these data form three training stages of approximately 36 million, 54 million, and 46 million examples, with separate Order and Click test sets from logged purchases and clicks.
We also evaluate on the Amazon Sports, Toys, and Clothing benchmarks \cite{he2016ups} and LastFM \cite{cantador2011hetrec}, following prior work \cite{rajput2023tiger,wan2026r3vae}.
Appendix~\ref{app:experimental-config} provides the dataset details.

\begin{table*}[!t]
\centering
\small
\setlength{\tabcolsep}{1.45pt}
\renewcommand{\arraystretch}{1.06}
\begin{tabular*}{\textwidth}{@{\extracolsep{\fill}}l*{11}{c}@{}}
\toprule
& \multicolumn{4}{c}{Downstream Evaluation} & \multicolumn{7}{c}{Codebook Quality}\\
\cmidrule(lr){2-5}\cmidrule(l){6-12}
Method & \multicolumn{2}{c}{Order} & \multicolumn{2}{c}{Click}
& \multicolumn{1}{c}{Full SID} & \multicolumn{3}{c}{Utilization (\%)} & \multicolumn{3}{c}{Gini}\\
\cmidrule(lr){2-3}\cmidrule(lr){4-5}\cmidrule(lr){6-6}\cmidrule(lr){7-9}\cmidrule(l){10-12}
& HitRate & MRR & HitRate & MRR & ICR (\%) & L1 & L2 & L3 & L1 & L2 & L3\\
\midrule
\multicolumn{12}{c}{\emph{Codebook: 1024--512--128}}\\
\cmidrule(lr){1-12}
RQ-VAE & 0.2474 & 0.0501 & 0.2921 & 0.0446 & 54.24 & \textbf{100.00} & 47.55 & 3.22 & 0.407 & 0.795 & 0.601\\
R3-VAE & 0.2289 & 0.0449 & 0.2788 & 0.0402 & 54.96 & 99.41 & \underline{51.04} & 3.34 & 0.427 & 0.783 & 0.592\\
RQ-KMeans & \underline{0.2913} & \underline{0.0644} & \underline{0.3279} & \underline{0.0540} & \underline{55.52} & \textbf{100.00} & 47.88 & \underline{3.61} & 0.413 & \underline{0.774} & \underline{0.569}\\
RQ-GMM & 0.2687 & 0.0594 & 0.3071 & 0.0483 & 50.66 & 99.90 & 38.57 & 3.06 & 0.412 & 0.783 & 0.598\\
PQ-KMeans & 0.0400 & 0.0061 & 0.0521 & 0.0054 & 50.45 & \textbf{100.00} & 22.52 & 0.67 & 0.351 & 0.934 & 0.891\\
OPQ-KMeans & 0.1242 & 0.0225 & 0.1393 & 0.0164 & 47.35 & \textbf{100.00} & 21.85 & 1.15 & \textbf{0.276} & 0.916 & 0.804\\
\midrule
PRQ-KMeans & \textbf{0.3128} & \textbf{0.0720} & \textbf{0.3488} & \textbf{0.0588} & \textbf{58.45} & \textbf{100.00} & \textbf{57.78} & \textbf{3.99} & \underline{0.298} & \textbf{0.758} & \textbf{0.546}\\
\midrule
\multicolumn{12}{c}{\emph{Codebook: 1024--512--128--64--64}}\\
\cmidrule(lr){1-12}
ResKmeansFSQ & 0.4038 & 0.1194 & 0.4364 & 0.0969 & 89.19 & \textbf{100.00} & \underline{47.88} & 3.61 & \underline{0.413} & \underline{0.774} & 0.569\\
RQ-OPQ & \underline{0.4276} & \underline{0.1340} & \underline{0.4650} & \underline{0.1083} & \textbf{98.99} & \textbf{100.00} & \underline{47.88} & \underline{8.00} & \underline{0.413} & \underline{0.774} & \underline{0.266}\\
PRQ-OPQ & \textbf{0.4359} & \textbf{0.1364} & \textbf{0.4713} & \textbf{0.1089} & \underline{98.74} & \textbf{100.00} & \textbf{57.78} & \textbf{8.42} & \textbf{0.298} & \textbf{0.758} & \textbf{0.241}\\
\bottomrule
\end{tabular*}
\normalfont\normalsize
  \caption{Codebook quality and downstream generative retrieval performance on the industrial dataset. ICR denotes the independent-code ratio, the percentage of occupied full SIDs assigned to exactly one item. For each codebook block, best and second-best values are bold and underlined, respectively; lower Gini is better.}
\label{tab:industry-main}
\end{table*}

\begin{table}[!t]
\centering
\scriptsize
\setlength{\tabcolsep}{0.4pt}
\renewcommand{\arraystretch}{1.08}
\begin{tabular*}{\columnwidth}{@{\extracolsep{\fill}}l*{8}{r}@{}}
\toprule
& \multicolumn{2}{c}{Sports} & \multicolumn{2}{c}{Toys} & \multicolumn{2}{c}{Clothing} & \multicolumn{2}{c}{LastFM}\\
\cmidrule(lr){2-3}\cmidrule(lr){4-5}\cmidrule(lr){6-7}\cmidrule(l){8-9}
Method & Recall & NDCG & Recall & NDCG & Recall & NDCG & Recall & NDCG\\
\midrule
RQ-VAE
& 0.0522 & \textbf{0.0219}
& 0.0751 & 0.0321
& 0.0357 & 0.0141
& 0.0082 & 0.0033\\
R3-VAE
& 0.0514 & 0.0211
& 0.0812 & \underline{0.0354}
& \underline{0.0361} & \underline{0.0147}
& 0.0074 & 0.0029\\
RQ-KMeans
& \underline{0.0528} & \underline{0.0215}
& \underline{0.0824} & 0.0346
& 0.0330 & 0.0132
& \underline{0.0115} & \underline{0.0042}\\
RQ-GMM
& 0.0506 & 0.0209
& 0.0711 & 0.0287
& 0.0341 & 0.0133
& 0.0082 & 0.0024\\
PQ-KMeans
& 0.0087 & 0.0033
& 0.0177 & 0.0080
& 0.0082 & 0.0034
& 0.0016 & 0.0007\\
OPQ-KMeans
& 0.0379 & 0.0151
& 0.0403 & 0.0171
& 0.0233 & 0.0087
& 0.0033 & 0.0014\\
\midrule
PRQ-KMeans
& \textbf{0.0531} & \textbf{0.0219}
& \textbf{0.0851} & \textbf{0.0361}
& \textbf{0.0382} & \textbf{0.0151}
& \textbf{0.0179} & \textbf{0.0076}\\
\bottomrule
\end{tabular*}
\normalfont\normalsize
\caption{Performance evaluation on the public benchmarks.}
\label{tab:public-main}
\end{table}

% \paragraph{Baselines.}
% We compare PRQ-KMeans with learned residual tokenizers RQ-VAE \cite{lee2022rqvae} and R3-VAE \cite{wan2026r3vae}, post-hoc tokenizers RQ-KMeans \cite{luo2024qarm} and RQ-GMM \cite{tong2026rqgmm}, and parallel quantizers PQ-KMeans \cite{jegou2011pq} and OPQ-KMeans \cite{ge2013opq}.
% For the five-level industrial setting, we also evaluate hybrid tokenizers that combine a hierarchical residual prefix with complementary suffix tokens \cite{chen2025onesearch,xia2026qarmv2}.
% RQ-OPQ \cite{chen2025onesearch} combines a three-token RQ-KMeans prefix with a two-token OPQ suffix, while ResKmeansFSQ \cite{xia2026qarmv2} combines the same prefix with a two-token FSQ suffix.
% PRQ-OPQ replaces the RQ-KMeans prefix with PRQ-KMeans while retaining the OPQ suffix as well as the third-layer balanced KMeans, which allows us to evaluate compatibility with the same industrial tokenizer interface.
% Appendix~\ref{app:experimental-config} provides the reproduction and alignment details.

\paragraph{Baselines.}
We compare PRQ-KMeans with learned residual tokenizers RQ-VAE \cite{lee2022rqvae} and R3-VAE \cite{wan2026r3vae}, post-hoc tokenizers RQ-KMeans \cite{luo2024qarm} and RQ-GMM \cite{tong2026rqgmm}, and parallel quantizers PQ-KMeans \cite{jegou2011pq} and OPQ-KMeans \cite{ge2013opq}. For the five-level industrial setting, we also evaluate hybrid tokenizers that combine a hierarchical residual prefix with suffix tokens \cite{chen2025onesearch,xia2026qarmv2}. RQ-OPQ \cite{chen2025onesearch} uses two RQ-KMeans levels, a balanced K-Means third level, and two OPQ suffix tokens. ResKmeansFSQ \cite{xia2026qarmv2} uses three RQ-KMeans levels followed by two FSQ suffix tokens. PRQ-OPQ replaces the first two RQ-KMeans levels of RQ-OPQ with PRQ-KMeans, while retaining its balanced third-level K-Means and
two OPQ suffix tokens. This construction evaluates the compatibility of PRQ-KMeans with the balanced third level and OPQ suffix used in RQ-OPQ. Appendix~\ref{app:experimental-config} provides the reproduction and alignment details.

\paragraph{Implementation and Evaluation.}
For the industrial dataset, we follow OneSearch \cite{chen2025onesearch}, using its 128-dimensional distilled-BGE query and item representations and BART-Base retriever \cite{lewis2020bart}.
All tokenizers share the three-stage training, decoding, and evaluation pipeline.
The three-level setting uses a 1024-512-128 codebook; the five-level setting uses the same three-level prefix followed by two 64-way suffix tokens.
For the public benchmarks, we use 768-dimensional Sentence-T5-Base item embeddings \cite{ni2022sentencet5}, a 256-256-256 codebook, and the TIGER pipeline \cite{rajput2023tiger,wan2026r3vae}.
PRQ-KMeans uses $k=2$ on the public benchmarks and $k=5$ on the industrial dataset due to their different codebook sizes, with $\beta=15$ in both settings.
Following OneSearch, industrial evaluation reports HitRate@50 and MRR@50 for Order and Click, together with the independent-code ratio (ICR), prefix utilization, and used-prefix Gini for codebook quality.
Public evaluation reports item-level Recall@20 and NDCG@20; Appendix~\ref{app:experimental-config} provides evaluation details.
The code will be released upon acceptance.

\subsection{Overall Performance}

\paragraph{Industrial benchmark.}
We compare three-level codebook quality (Table~\ref{tab:industry-main}).
Among all evaluated methods, PRQ-KMeans obtains the highest ICR and L2--L3 prefix utilization together with the lowest L2--L3 Gini.
Relative to RQ-KMeans, ICR increases from 55.52\% to 58.45\%, L2--L3 utilization rises from 47.88\%/3.61\% to 57.78\%/3.99\%, and L2--L3 Gini decreases from 0.774/0.569 to 0.758/0.546.

PRQ-KMeans also improves downstream generative retrieval.
RQ-KMeans is the strongest three-level baseline on all four metrics.
Relative to RQ-KMeans, PRQ-KMeans improves Order HitRate and MRR by 7.4\% and 11.8\%, respectively, and Click HitRate and MRR by 6.4\% and 8.9\%.
In the five-level hybrid setting, PRQ-OPQ achieves the best Order and Click results on all four downstream metrics among the three aligned tokenizers and improves over RQ-OPQ by 0.6\%--1.9\%.
This result supports the compatibility of the PRQ-KMeans prefix with an OPQ suffix under the same industrial tokenizer interface.

% EVIDENCE_STATUS [INDUSTRIAL_3L]: Observed fixed-run downstream and
% codebook-quality values transcribed from results/kuaishou/expierment.md.
% Full resolved configs, logs, fixed seeds, and evaluator artifacts remain
% provenance gates before the result is upgraded from Observed.
% EVIDENCE_STATUS [INDUSTRIAL_5L]: Author-reported fixed-run downstream and
% codebook aggregates recorded in results/kuaishou/expierment.md. Run IDs,
% resolved configs, fixed seeds, logs, SID exports, and evaluator artifacts
% remain provenance gates before these results are upgraded from Observed.

\paragraph{Public benchmarks.}
Across four benchmarks, PRQ-KMeans obtains the best or tied-best value on all eight metrics and exceeds RQ-KMeans in every comparison (Table~\ref{tab:public-main}).
The gains are modest on Sports and Toys, larger on Clothing, and most pronounced on LastFM, where Recall increases from 0.0115 to 0.0179 and NDCG from 0.0042 to 0.0076 relative to RQ-KMeans.
These results extend the evidence beyond the industrial setting to multiple public domains.

\begin{table}[!t]
\centering
\small
\setlength{\tabcolsep}{4.0pt}
\begin{tabular}{@{}lcccc@{}}
\toprule
Variant & \multicolumn{2}{c}{Order} & \multicolumn{2}{c}{Click}\\
\cmidrule(lr){2-3}\cmidrule(l){4-5}
& HitRate & MRR & HitRate & MRR\\
\midrule
PRQ-KMeans & \textbf{0.3128} & \textbf{0.0720} & \textbf{0.3488} & \textbf{0.0588}\\
$-$ Global & 0.3017 & 0.0697 & 0.3379 & 0.0560\\
$-$ Soft & \underline{0.3067} & \underline{0.0705} & \underline{0.3460} & \underline{0.0584}\\
$-$ Projection & 0.2985 & 0.0687 & 0.3328 & 0.0548\\
\bottomrule
\end{tabular}
\normalfont\normalsize
\caption{Ablation study on the industrial dataset.}
\label{tab:ablation-kuaishou}
\end{table}

% EVIDENCE_STATUS [PUBLIC_MAIN]: Baseline rows are transcribed from
% notes/results_tiger_item_level_final.md and the aligned R3 export recorded
% in notes/public_tiger_experiments.md. The PRQ-KMeans row uses the paired
% full-method endpoint registered in results/public/ablation_results.md.
% Before paper-ready release, add test sample counts and close the
% evaluator/config provenance audit.

\subsection{Ablation Study}
\label{sec:component-ablations}

Tables~\ref{tab:ablation-kuaishou} and~\ref{tab:ablation-public} (in Appendix) report the industrial and public ablations, respectively.
We compare the complete model with three variants: \emph{$-$ Global} omits global component removal; \emph{$-$ Soft} replaces Top-$k$ soft refinement with hard centroid updates; and \emph{$-$ Projection} retains Global and Top-$k$ refinement but replaces cosine fitting and assignment with Euclidean counterparts and projection residualization with full-centroid subtraction.
The complete PRQ-KMeans model achieves the best result on every industrial metric and the best or tied-best result on all eight public metrics, tying \emph{$-$ Global} on Clothing NDCG.
On the industrial dataset, removing Projection produces the largest decrease on all four metrics, while removing Global or Soft also yields consistent drops.
On the public benchmarks, removing Projection produces the largest decrease on all six Amazon metrics, while LastFM is most sensitive to Soft.
Removing Global lowers seven of eight metrics and ties the complete model on Clothing NDCG.

\subsection{Sensitivity and Robustness}
\paragraph{Hyperparameter sensitivity.}

% We examine sensitivity to the Top-$k$ neighborhood size $k$ and soft-weight concentration $\beta$, which control the candidate count and weight sharpness, respectively.
% Figure~\ref{fig:hyperparameter-sensitivity} reports sensitivity on the industrial dataset under the same downstream protocol as Table~\ref{tab:industry-main}.
% At fixed $\beta=15$, every soft-refinement setting with $k\ge2$ yields higher Order and Click HitRate than the hard-update setting ($k=1$).
% At fixed $k=5$, $\beta=5$ gives the lowest HitRate on both tasks, while $\beta\ge15$ yields broadly comparable results, with no consistent benefit from further increasing $\beta$.
% Appendix~\ref{app:sensitivity} provides complementary sensitivity results on the four public benchmarks.

\begin{figure}[t]
  \centering
  \begin{subfigure}[t]{0.485\columnwidth}
    \centering
    \includegraphics[width=\linewidth]{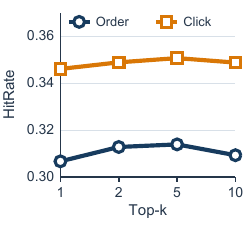}
    \caption{Top-$k$}
    \label{fig:topk-sensitivity}
  \end{subfigure}\hfill
  \begin{subfigure}[t]{0.485\columnwidth}
    \centering
    \includegraphics[width=\linewidth]{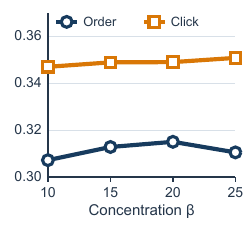}
    \caption{Weight concentration}
    \label{fig:beta-sensitivity}
  \end{subfigure}
  \caption{Hyperparameter sensitivity on the industrial dataset.}
  \label{fig:hyperparameter-sensitivity}
\end{figure}

We study sensitivity to $k$, the number of nearby centroids receiving weighted updates, and $\beta$, which controls weight concentration.
Figure~\ref{fig:hyperparameter-sensitivity} shows that $k=2$ and $k=5$ perform better than $k=1$ and $k=10$.
This suggests a trade-off: a small neighborhood limits sharing near cluster boundaries, while larger neighborhoods may weaken locality by updating less similar centroids.
For $\beta$, $\beta=10$ performs worse, while results remain comparable across $\beta=15$--25.
This suggests that diffuse weights may obscure similarity differences among the centroids, whereas sharper weights better prioritize the closest candidates.
Appendix~\ref{app:sensitivity} provides results on the public benchmarks.

\paragraph{Embedding robustness.}

\begin{figure}[t]
  \centering
  \begin{subfigure}[t]{0.485\columnwidth}
    \centering
    \includegraphics[width=\linewidth]{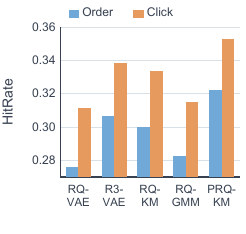}
    \caption{HitRate}
  \end{subfigure}\hfill
  \begin{subfigure}[t]{0.485\columnwidth}
    \centering
    \includegraphics[width=\linewidth]{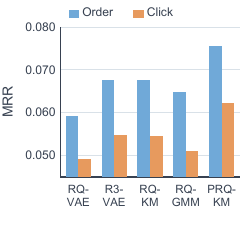}
    \caption{MRR}
  \end{subfigure}
  \caption{Embedding robustness on the industrial dataset.
  The aligned BGE representations used in the main experiment are replaced with aligned Qwen3 representations, while the codebook sizes and downstream protocol remain unchanged.
  KM denotes K-Means.}
  \label{fig:embedding-robustness}
\end{figure}

To evaluate robustness across input embeddings, we replace the aligned BGE representations used in the main experiment with aligned Qwen3-Embedding-0.6B representations \cite{zhang2025qwen3embedding}.
We rerun five tokenizers using these 128-dimensional representations, with the alignment procedure, codebook sizes, and downstream protocol unchanged.
Figure~\ref{fig:embedding-robustness} shows that PRQ-KMeans achieves the highest HitRate and MRR on both Order and Click under the alternative embedding, supporting its robustness to different input embeddings.

\subsection{Qualitative Analysis}

\paragraph{Centroid visualization.}
To visualize how residual construction shapes the learned codebooks, we use the L1--L3 centroids of RQ-KMeans and PRQ-KMeans and jointly embed them in two dimensions with t-SNE \cite{vandermaaten2008visualizing}.
In RQ-KMeans, L1 centroids span a broad region, whereas L2 and especially L3 centroids form a dense core.
This pattern is consistent with residual carryover: when the selected-centroid component remains, later codebooks receive representations that still vary along earlier centroid directions and may revisit previously represented variation.
In contrast, PRQ-KMeans removes this component before fitting the next codebook, and its L2 and L3 centroids cover a broader region.
Together with the higher later-level utilization and lower Gini in Table~\ref{tab:industry-main}, this visualization provides qualitative support for progressive commonality removal.

\begin{figure}[t]
  \centering
  \includegraphics[width=\columnwidth]{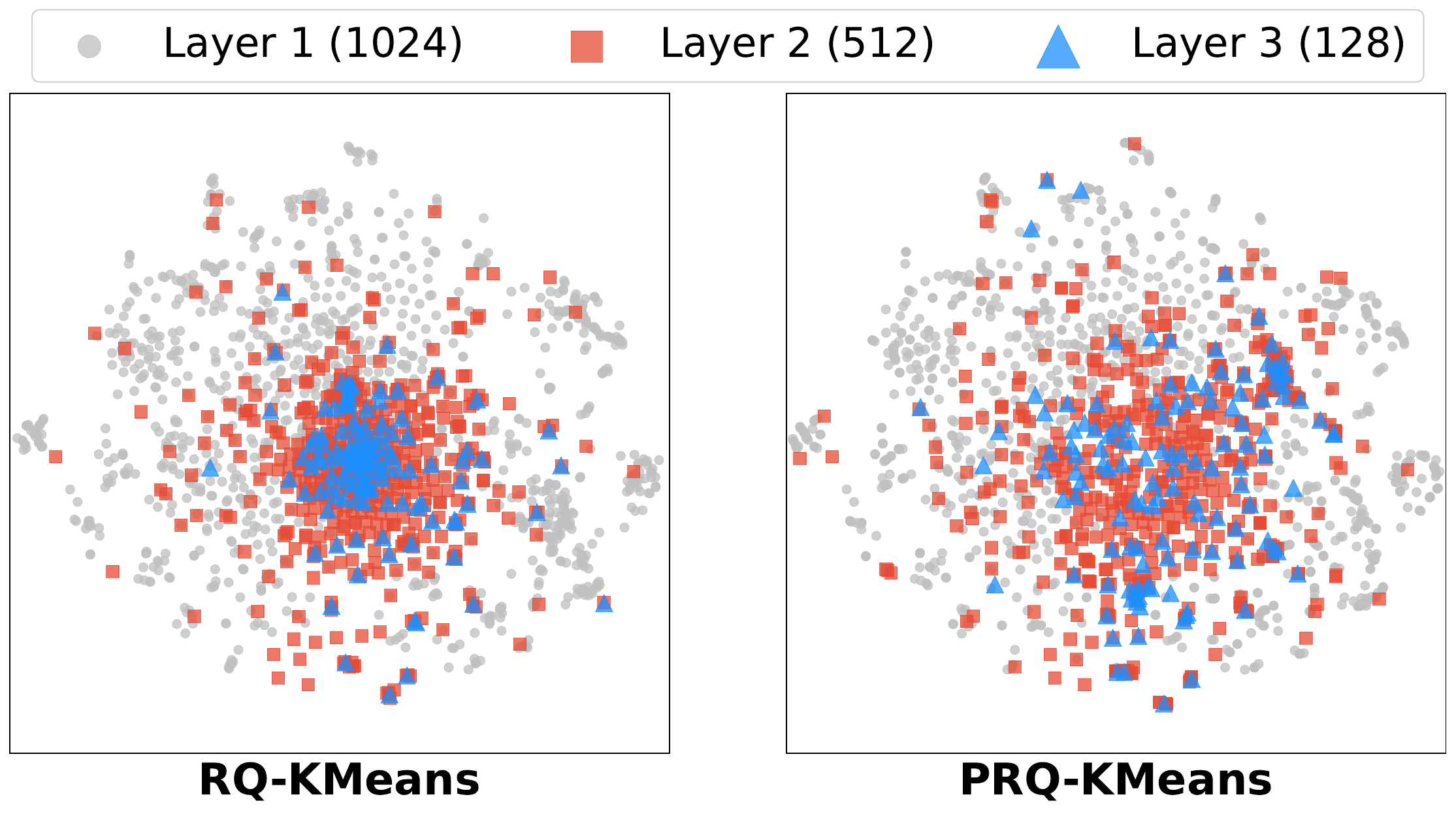}
  \caption{Centroid visualization on the industrial dataset. L1--L3 centroids from RQ-KMeans and PRQ-KMeans are jointly embedded using one t-SNE fit; both panels share coordinates and axis limits, and colors and markers denote SID levels.}
  \label{fig:tsne-centroids}
\end{figure}

\paragraph{Case study.}
We trace 12 Kitty-themed slippers assigned to one RQ-KMeans full SID.
Under PRQ-KMeans, these items retain one shared level-1 prefix, separate into two level-2 prefixes, and form three full SIDs containing four items each.
In this group, the later tokens distinguish observed indoor soft-soled, outdoor/dormitory, and mixed-use variants while retaining the shared product theme.
Appendix~\ref{app:qualitative-prefix} reports the exact collision-group statistics and selection procedure; it also presents a case study of the weak-intent query ``hotel.''

% Keep the result tables ahead of the conclusion without forcing a float-only page.
\section{Conclusion}

This paper presents PRQ-KMeans, a post-hoc hierarchical SID tokenizer based on progressive commonality removal. It removes the global component before the first level and refines each codebook with Top-$k$ soft weights. After assignment, it removes the selected-centroid component through projection before fitting the next codebook. PRQ-KMeans achieves the best overall downstream retrieval performance on the industrial dataset and four public benchmarks. Component ablations support the complete design. Together, these results support explicitly controlling what is passed between successive SID levels.

\bibliography{references}

@article{tay2022dsi,
  title={Transformer Memory as a Differentiable Search Index},
  author={Tay, Yi and Tran, Vinh and Dehghani, Mostafa and Ni, Jianmo and Bahri, Dara and Mehta, Harsh and Qin, Zhen and Hui, Kai and Zhao, Zhe and Gupta, Jai and others},
  journal={Advances in Neural Information Processing Systems},
  volume={35},
  pages={21831--21843},
  year={2022}
}

@article{wang2022nci,
  title={A neural corpus indexer for document retrieval},
  author={Wang, Yujing and Hou, Yingyan and Wang, Haonan and Miao, Ziming and Wu, Shibin and Chen, Qi and Xia, Yuqing and Chi, Chengmin and Zhao, Guoshuai and Liu, Zheng and others},
  journal={Advances in Neural Information Processing Systems},
  volume={35},
  pages={25600--25614},
  year={2022}
}

@article{sun2023genret,
  title={Learning to tokenize for generative retrieval},
  author={Sun, Weiwei and Yan, Lingyong and Chen, Zheng and Wang, Shuaiqiang and Zhu, Haichao and Ren, Pengjie and Chen, Zhumin and Yin, Dawei and Rijke, Maarten and Ren, Zhaochun},
  journal={Advances in Neural Information Processing Systems},
  volume={36},
  pages={46345--46361},
  year={2023}
}

@inproceedings{lee2022rqvae,
  title={Autoregressive image generation using residual quantization},
  author={Lee, Doyup and Kim, Chiheon and Kim, Saehoon and Cho, Minsu and Han, Wook-Shin},
  booktitle={Proceedings of the IEEE/CVF conference on computer vision and pattern recognition},
  pages={11523--11532},
  year={2022}
}

@inproceedings{rajput2023tiger,
  title={Recommender systems with generative retrieval},
  author={Rajput, Shashank and Mehta, Nikhil and Singh, Anima and Keshavan, Raghunandan Hulikal and Vu, Trung and Heldt, Lukasz and Hong, Lichan and Tay, Yi and Tran, Vinh Q and Samost, Jonah and others},
  booktitle={Thirty-seventh Conference on Neural Information Processing Systems},
  year={2023}
}

@inproceedings{he2016ups,
  title={Ups and downs: Modeling the visual evolution of fashion trends with one-class collaborative filtering},
  author={He, Ruining and McAuley, Julian},
  booktitle={proceedings of the 25th international conference on world wide web},
  pages={507--517},
  year={2016}
}

@inproceedings{cantador2011hetrec,
  title={Second workshop on information heterogeneity and fusion in recommender systems (HetRec2011)},
  author={Cantador, Iv{\'a}n and Brusilovsky, Peter and Kuflik, Tsvi},
  booktitle={Proceedings of the fifth ACM conference on Recommender systems},
  pages={387--388},
  year={2011}
}

@inproceedings{singh2023semanticids,
  title={Better generalization with semantic ids: A case study in ranking for recommendations},
  author={Singh, Anima and Vu, Trung and Mehta, Nikhil and Keshavan, Raghunandan and Sathiamoorthy, Maheswaran and Zheng, Yilin and Hong, Lichan and Heldt, Lukasz and Wei, Li and Tandon, Devansh and others},
  booktitle={Proceedings of the 18th ACM Conference on Recommender Systems},
  pages={1039--1044},
  year={2024}
}

@inproceedings{luo2024qarm,
  title={Qarm: Quantitative alignment multi-modal recommendation at kuaishou},
  author={Luo, Xinchen and Cao, Jiangxia and Sun, Tianyu and Yu, Jinkai and Huang, Rui and Yuan, Wei and Lin, Hezheng and Zheng, Yichen and Wang, Shiyao and Hu, Qigen and others},
  booktitle={Proceedings of the 34th ACM International Conference on Information and Knowledge Management},
  pages={5915--5922},
  year={2025}
}

@inproceedings{wang2024letter,
  title={Learnable item tokenization for generative recommendation},
  author={Wang, Wenjie and Bao, Honghui and Lin, Xinyu and Zhang, Jizhi and Li, Yongqi and Feng, Fuli and Ng, See-Kiong and Chua, Tat-Seng},
  booktitle={Proceedings of the 33rd ACM International Conference on Information and Knowledge Management},
  pages={2400--2409},
  year={2024}
}

@inproceedings{zhu2024cost,
  title={Cost: Contrastive quantization based semantic tokenization for generative recommendation},
  author={Zhu, Jieming and Jin, Mengqun and Liu, Qijiong and Qiu, Zexuan and Dong, Zhenhua and Li, Xiu},
  booktitle={Proceedings of the 18th ACM Conference on Recommender Systems},
  pages={969--974},
  year={2024}
}

@article{liu2025etegrec,
  title={Generative recommender with end-to-end learnable item tokenization},
  author={Liu, Enze and Zheng, Bowen and Ling, Cheng and Hu, Lantao and Li, Han and Zhao, Wayne Xin},
  journal={arXiv preprint arXiv:2409.05546},
  year={2024}
}

@article{wan2026r3vae,
  title={R3-VAE: Reference Vector-Guided Rating Residual Quantization VAE for Generative Recommendation},
  author={Wan, Qiang and Yang, Ze and Yang, Dawei and Fan, Ying and Yan, Xin and Liu, Siyang and Liu, Yicong and Zhang, Chenwei and Xu, Wei and Qin, Jiahao and others},
  journal={arXiv preprint arXiv:2604.11440},
  year={2026}
}

@inproceedings{yin2026dos,
  title={DOS: Dual-Flow Orthogonal Semantic IDs for Recommendation in Meituan},
  author={Yin, Junwei and Kou, Senjie and Li, Changhao and Wang, Shuli and Huang, Yinqiu and Wei, Xue and Zhu, Yinhua and Wang, Haitao and Wang, Xingxing},
  booktitle={Proceedings of the ACM Web Conference 2026},
  pages={8305--8308},
  year={2026}
}

@article{zhang2026reliable,
  title={How Reliable Are Semantic-ID Tokenizer Comparisons in Generative Recommendation?},
  author={Zhang, Qian and Szymanski, Lech and Zhang, Haibo and Deng, Jeremiah D},
  journal={arXiv preprint arXiv:2605.25330},
  year={2026}
}

@article{ni2022sentencet5,
  title={Sentence-t5: Scalable sentence encoders from pre-trained text-to-text models},
  author={Ni, Jianmo and Abrego, Gustavo Hernandez and Constant, Noah and Ma, Ji and Hall, Keith B and Cer, Daniel and Yang, Yinfei},
  journal={arXiv preprint arXiv:2108.08877},
  year={2021}
}

@inproceedings{lewis2020bart,
  title={BART: Denoising sequence-to-sequence pre-training for natural language generation, translation, and comprehension},
  author={Lewis, Mike and Liu, Yinhan and Goyal, Naman and Ghazvininejad, Marjan and Mohamed, Abdelrahman and Levy, Omer and Stoyanov, Veselin and Zettlemoyer, Luke},
  booktitle={Proceedings of the 58th annual meeting of the association for computational linguistics},
  pages={7871--7880},
  year={2020}
}

@article{jegou2011pq,
  title={Product quantization for nearest neighbor search},
  author={Jegou, Herve and Douze, Matthijs and Schmid, Cordelia},
  journal={IEEE Transactions on Pattern Analysis \& Machine Intelligence},
  volume={33},
  number={01},
  pages={117--128},
  year={2011},
  publisher={IEEE Computer Society}
}

@inproceedings{ge2013opq,
  title={Optimized product quantization for approximate nearest neighbor search},
  author={Ge, Tiezheng and He, Kaiming and Ke, Qifa and Sun, Jian},
  booktitle={Proceedings of the IEEE conference on computer vision and pattern recognition},
  pages={2946--2953},
  year={2013}
}

@article{shi2025lladarec,
  title={LLaDA-Rec: Discrete Diffusion for Parallel Semantic ID Generation in Generative Recommendation},
  author={Shi, Teng and Shen, Chenglei and Yu, Weijie and Nie, Shen and Li, Chongxuan and Zhang, Xiao and He, Ming and Han, Yan and Xu, Jun},
  journal={arXiv preprint arXiv:2511.06254},
  year={2025}
}

@article{tong2026rqgmm,
  title={RQ-GMM: Residual Quantized Gaussian Mixture Model for Multimodal Semantic Discretization in CTR Prediction},
  author={Tong, Ziye and Liu, Jiahao and Zhang, Weimin and Ruan, Hongji and Tang, Derick and Zeng, Zhanpeng and Zeng, Qinsong and Zhang, Peng and Lu, Tun and Gu, Ning},
  journal={arXiv preprint arXiv:2602.12593},
  year={2026}
}

@article{xia2026qarmv2,
  title={QARM V2: Quantitative Alignment Multi-Modal Recommendation for Reasoning User Sequence Modeling},
  author={Xia, Tian and Zhang, Jiaqi and Liu, Yueyang and Dou, Hongjian and Yin, Tingya and Cao, Jiangxia and Liang, Xulei and Xie, Tianlu and Liu, Lihao and Chen, Xiang and others},
  journal={arXiv preprint arXiv:2602.08559},
  year={2026}
}

@article{choi2026quantizing,
  title={Quantizing Intent: Cross-Domain Semantic IDs from Organic Activity for Industrial Ranking},
  author={Choi, Julie and Ye, Haoran and Ding, Zhiwei and Long, Bo and Zelditch, Benjamin and Vats, Arpita},
  journal={arXiv preprint arXiv:2606.01396},
  year={2026}
}

@article{chen2025onesearch,
  title={Onesearch: A preliminary exploration of the unified end-to-end generative framework for e-commerce search},
  author={Chen, Ben and Guo, Xian and Wang, Siyuan and Liang, Zihan and Lv, Yue and Ma, Yufei and Xiao, Xinlong and Xue, Bowen and Zhang, Xuxin and Yang, Ying and others},
  journal={arXiv preprint arXiv:2509.03236},
  year={2025}
}

@article{li2026kuaisearch,
  title={KuaiSearch: A Large-Scale E-Commerce Search Dataset for Recall, Ranking, and Relevance},
  author={Li, Yupeng and Chen, Ben and Cheng, Mingyue and Liu, Zhiding and Zhang, Xuxin and Lei, Chenyi and Ou, Wenwu},
  journal={arXiv preprint arXiv:2602.11518},
  year={2026}
}

@article{deng2025onerec,
  title={Onerec: Unifying retrieve and rank with generative recommender and iterative preference alignment},
  author={Deng, Jiaxin and Wang, Shiyao and Cai, Kuo and Ren, Lejian and Hu, Qigen and Ding, Weifeng and Luo, Qiang and Zhou, Guorui},
  journal={arXiv preprint arXiv:2502.18965},
  year={2025}
}

@inproceedings{lee2023glen,
  title={GLEN: Generative retrieval via lexical index learning},
  author={Lee, Sunkyung and Choi, Minjin and Lee, Jongwuk},
  booktitle={Proceedings of the 2023 Conference on Empirical Methods in Natural Language Processing},
  pages={7693--7704},
  year={2023}
}

@inproceedings{zhang2025merge,
  title={Multi-level Relevance Document Identifier Learning for Generative Retrieval},
  author={Zhang, Fuwei and Liu, Xiaoyu and Jia, Xinyu and Zhang, Yingfei and Zhang, Shuai and Li, Xiang and Zhuang, Fuzhen and Lin, Wei and Zhang, Zhao},
  booktitle={Proceedings of the 63rd Annual Meeting of the Association for Computational Linguistics (Volume 1: Long Papers)},
  pages={10066--10080},
  year={2025}
}

@article{fu2026diger,
  title={Differentiable Semantic ID for Generative Recommendation},
  author={Fu, Junchen and Ge, Xuri and Karatzoglou, Alexandros and Arapakis, Ioannis and Verberne, Suzan and Jose, Joemon M and Ren, Zhaochun},
  journal={arXiv preprint arXiv:2601.19711},
  year={2026}
}

@article{jiang2026unisid,
  title={End-to-End Semantic ID Generation for Generative Advertisement Recommendation},
  author={Jiang, Jie and Zhang, Xinxun and Zhang, Enming and Xiong, Yuling and Zhang, Jun and Wang, Jingwen and Yu, Huan and Wang, Yuxiang and Wang, Hao and Yan, Xiao and others},
  journal={arXiv preprint arXiv:2602.10445},
  year={2026}
}

@article{xia2026acerec,
  title={Unleash the Potential of Long Semantic IDs for Generative Recommendation},
  author={Xia, Ming and Zhou, Zhiqin and Ma, Guoxin and Huang, Dongmin},
  journal={arXiv preprint arXiv:2602.13573},
  year={2026}
}

@article{liang2026resid,
  title={Rethinking Generative Recommender Tokenizer: Recsys-Native Encoding and Semantic Quantization Beyond LLMs},
  author={Liang, Yu and Zhang, Zhongjin and Zhu, Yuxuan and Zhang, Kerui and Guo, Zhiluohan and Zhou, Wenhang and Yang, Zonqi and Wu, Kangle and Ni, Yabo and Zeng, Anxiang and others},
  journal={arXiv preprint arXiv:2602.02338},
  year={2026}
}

@inproceedings{wang2026hhq,
  title={Distilling Large Embeddings via Hyperspherical Householder Quantization},
  author={Wang, Yihang and Wu, Bin and Su, Yueyang and Zhang, Tianfu and Du, Yiqi and Yu, Lei and Guo, Jiafeng and Cheng, Xueqi},
  booktitle={Proceedings of the 64th Annual Meeting of the Association for Computational Linguistics (Volume 1: Long Papers)},
  pages={10562--10576},
  year={2026}
}

@article{chen2026onesearchv2,
  title={OneSearch-V2: The Latent Reasoning Enhanced Self-distillation Generative Search Framework},
  author={Chen, Ben and Wang, Siyuan and Ma, Yufei and Liang, Zihan and Zhang, Xuxin and Lv, Yue and Yang, Ying and Dai, Huangyu and Mao, Lingtao and Zhao, Tong and others},
  journal={arXiv preprint arXiv:2603.24422},
  year={2026}
}

@article{zhang2025qwen3embedding,
  title={Qwen3 embedding: Advancing text embedding and reranking through foundation models},
  author={Zhang, Yanzhao and Li, Mingxin and Long, Dingkun and Zhang, Xin and Lin, Huan and Yang, Baosong and Xie, Pengjun and Yang, An and Liu, Dayiheng and Lin, Junyang and others},
  journal={arXiv preprint arXiv:2506.05176},
  year={2025}
}

@article{vandermaaten2008visualizing,
  title={Visualizing data using t-SNE.},
  author={Van der Maaten, Laurens and Hinton, Geoffrey},
  journal={Journal of machine learning research},
  volume={9},
  number={11},
  year={2008}
}
\clearpage
\appendix
\setcounter{figure}{0}
\renewcommand{\thefigure}{A\arabic{figure}}
\setcounter{table}{0}
\renewcommand{\thetable}{A\arabic{table}}

\section{Theoretical Proofs}
\label{app:proofs}

We first derive how representation-level carryover behaves within a cluster,
including the case in which the centroid is the arithmetic mean of its
assigned representations. We then prove the geometric properties of the
projection residual used by PRQ-KMeans.

\subsection{Residual-Carryover Identity}

Consider a nonzero centroid $\vc$, let $\rho=\norm{\vc}$ and
$\hat{\vc}=\vc/\rho$, and fix a nonempty assigned set $\mathcal{I}_j$.
For each $i\in\mathcal{I}_j$, write
$\vr_i=\alpha_i\hat{\vc}+\vu_i$, where
$\alpha_i=\vr_i^\top\hat{\vc}$ and $\vu_i\perp\hat{\vc}$, and let
$\bar\alpha_j=|\mathcal{I}_j|^{-1}
\sum_{i\in\mathcal{I}_j}\alpha_i$. Because the carryover
coefficient is $\alpha_i-\rho$, its mean squared magnitude satisfies
\begin{equation}
\frac{1}{|\mathcal{I}_j|}
\sum_{i\in\mathcal{I}_j}(\alpha_i-\rho)^2
=
\operatorname{Var}_{\mathcal{I}_j}(\alpha)
+(\bar\alpha_j-\rho)^2.
\label{eq:cluster-coefficient-variance}
\end{equation}
Here,
$\operatorname{Var}_{\mathcal{I}_j}(\alpha)
=|\mathcal{I}_j|^{-1}
\sum_{i\in\mathcal{I}_j}(\alpha_i-\bar\alpha_j)^2$
denotes the population variance within the assigned cluster. When $\vc$ is
the arithmetic mean of the assigned representations,
$\bar\alpha_j=\hat{\vc}^{\top}\vc=\rho$. The mean squared carryover then
equals the within-cluster variance of $\alpha_i$ and becomes zero only when
all assigned representations have the same coefficient along the centroid
direction. For a non-mean centroid, the second term in
Eq.~\eqref{eq:cluster-coefficient-variance} captures the additional offset.

\subsection{Proof of Proposition~\ref{prop:projection-properties}}

The identity above characterizes why full-centroid subtraction can retain a
component along the selected centroid. We now prove that the projection
residual removes this component with the minimum Euclidean change, as stated
in Proposition~\ref{prop:projection-properties}.
Let $\vc\in\mathbb{R}^{d}$ be nonzero and define
\begin{equation}
\vv^\star
=\vr-\frac{\vr^\top\vc}{\norm{\vc}^{2}}\vc.
\end{equation}

\paragraph{Exact orthogonality.}

By direct expansion,
\begin{equation}
\vc^\top\vv^\star
=
\vc^\top\vr-
\frac{\vr^\top\vc}{\norm{\vc}^{2}}\norm{\vc}^{2}
=0.
\end{equation}
If $\vv^\star\neq\vzero$, normalization multiplies it by a positive scalar
and therefore preserves orthogonality.

\paragraph{Minimum Euclidean change.}

The feasible set $\mathcal{S}=\{\vv:\vv^\top\vc=0\}$ is a closed linear subspace.
Because $\vv^\star\in\mathcal{S}$ and
$\vv^\star-\vr=-(\vr^\top\vc/\norm{\vc}^{2})\vc
\in\operatorname{span}(\vc)=\mathcal{S}^{\perp}$, any feasible $\vv\in\mathcal{S}$
satisfies
\begin{equation}
\norm{\vv-\vr}^{2}
=
\norm{\vv-\vv^\star}^{2}
+\norm{\vv^\star-\vr}^{2}.
\end{equation}
The second term is fixed and the first is nonnegative, with equality only
when $\vv=\vv^\star$. Thus $\vv^\star$ is the unique solution to
Eq.~\eqref{eq:orthogonal-residual-opt}. This minimum-change property applies
to the projection residual before normalization.

\section{Algorithm Details}
\label{app:algorithm}

Algorithm~\ref{alg:prq-kmeans} presents the complete fitting and encoding
procedure. The fitting stage learns the global mean and level-wise codebooks
from the input embeddings. The encoding stage freezes these parameters and
generates a hierarchical SID for each embedding.

\begin{algorithm}[!t]
\caption{PRQ-KMeans fitting and SID encoding}
\label{alg:prq-kmeans}
\small
\textbf{Input}: embeddings $\{\vx_i\}_{i=1}^{N}$, depth $L$, codebook sizes
$\{K_\ell\}_{\ell=1}^{L}$, neighborhood size
$k$, concentration $\beta$, and the number of refinement iterations\\
\textbf{Output}: global mean $\vmu$, codebooks
$\{\mathcal{C}^{(\ell)}\}_{\ell=1}^{L}$, and hierarchical SIDs
\begin{algorithmic}
\STATE \textbf{Tokenizer Fitting}
\STATE \textit{Global Component Removal}
\STATE $\bar{\vx}_i\leftarrow\normalize{\vx_i}$ for all $i$
\STATE $\vmu \leftarrow N^{-1}\sum_i\bar{\vx}_i$
\STATE $\vr_i^{(1)}\leftarrow
  \normalize{\bar{\vx}_i-
  (\bar{\vx}_i^\top\vmu/\norm{\vmu}^{2})\vmu}$ for all $i$
\FOR{$\ell=1,\ldots,L$}
  \STATE Sample $K_\ell$ current residuals to initialize the centroids
  \STATE \textit{Top-$k$ Soft Centroid Refinement}
  \FOR{each prescribed refinement iteration}
    \STATE $s_{ij}^{(\ell)}\leftarrow
      \operatorname{cos}(\vr_i^{(\ell)},\vc_j^{(\ell)})$
    \STATE $\mathcal{N}_{k,i}^{(\ell)}\leftarrow
      \topk_j s_{ij}^{(\ell)}$
    \STATE $\displaystyle
      w_{ij}^{(\ell)}\leftarrow
      \frac{\exp(\beta s_{ij}^{(\ell)})}
      {\sum_{t\in\mathcal{N}_{k,i}^{(\ell)}}
      \exp(\beta s_{it}^{(\ell)})}$
      for $j\in\mathcal{N}_{k,i}^{(\ell)}$; set it to zero otherwise
    \STATE $\displaystyle
      \vc_j^{(\ell)}\leftarrow
      \frac{\sum_i w_{ij}^{(\ell)}\vr_i^{(\ell)}}
      {\sum_i w_{ij}^{(\ell)}}$
  \ENDFOR
  \STATE \textit{Hard Assignment}
  \STATE $z_i^{(\ell)}\leftarrow
    \arg\max_j\operatorname{cos}(\vr_i^{(\ell)},\vc_j^{(\ell)})$
  \STATE $\vc_i^{(\ell)}\leftarrow
    \vc_{z_i^{(\ell)}}^{(\ell)}$
  \IF{$\ell<L$}
    \STATE \textit{Projection Residual}
    \STATE $\vr_i^{(\ell+1)}\leftarrow
      \normalize{\vr_i^{(\ell)}
      -\frac{\vr_i^{(\ell)\top}\vc_i^{(\ell)}}
      {\norm{\vc_i^{(\ell)}}^{2}}\vc_i^{(\ell)}}$
  \ENDIF
\ENDFOR
\STATE \textbf{SID Encoding}
\FOR{each input embedding $\vx_i$}
  \STATE $\bar{\vx}_i\leftarrow\normalize{\vx_i}$
  \STATE $\vr_i^{(1)}\leftarrow
    \normalize{\bar{\vx}_i-
    (\bar{\vx}_i^\top\vmu/\norm{\vmu}^{2})\vmu}$
  \FOR{$\ell=1,\ldots,L$}
    \STATE $z_i^{(\ell)}\leftarrow
      \arg\max_j\operatorname{cos}(\vr_i^{(\ell)},\vc_j^{(\ell)})$
    \STATE $\vc_i^{(\ell)}\leftarrow
      \vc_{z_i^{(\ell)}}^{(\ell)}$
    \IF{$\ell<L$}
      \STATE $\vr_i^{(\ell+1)}\leftarrow
        \normalize{\vr_i^{(\ell)}
        -\frac{\vr_i^{(\ell)\top}\vc_i^{(\ell)}}
        {\norm{\vc_i^{(\ell)}}^{2}}\vc_i^{(\ell)}}$
    \ENDIF
  \ENDFOR
  \STATE Store $(z_i^{(1)},\ldots,z_i^{(L)})$ as the SID of $\vx_i$
\ENDFOR
\STATE \textbf{return} $\vmu$, $\{\mathcal{C}^{(\ell)}\}_{\ell=1}^{L}$,
  and all stored SIDs
\end{algorithmic}
\end{algorithm}

\section{Experimental Details}
\label{app:experimental-config}

This section describes the public and industrial benchmark protocols, the
baseline configurations, and the metrics and reproducibility
settings used throughout the experiments.

\subsection{Public Benchmark Setup}

The public evaluation uses the GenRec data processing and TIGER
encoder--decoder pipeline \cite{rajput2023tiger,wan2026r3vae} on Sports,
Toys, Clothing, and LastFM. Tokenizers are fitted on 768-dimensional
Sentence-T5-Base item representations \cite{ni2022sentencet5}, and every
tokenizer produces a three-level SID with a 256-256-256 codebook.
PRQ-KMeans uses $k=2$ and $\beta=15$.

All SIDs are evaluated with the same TIGER encoder--decoder, which has four
encoder layers, four decoder layers, model dimension 128, feed-forward
dimension 1024, and six attention heads. Training runs for at most 200
epochs with learning rate $10^{-4}$, zero weight decay, training batch size
256, and inference batch size 96. The best validation checkpoint is selected
with patience 10, and decoding uses beam size 30. The public results use
item-level Recall and NDCG. While some previous work reports SID-level
metrics, we use item-level Recall and NDCG to account for SID collisions
\cite{zhang2026reliable}. Specifically, for each predicted SID with multiple
mapped items, the evaluator randomly selects one representative item with
fixed seed 42; duplicate item predictions are removed before metric computation.

\subsection{Industrial Benchmark Setup}

The industrial benchmark follows the OneSearch training and evaluation
framework \cite{chen2025onesearch}.
Tokenizer fitting uses 7,797,542 items and 9,046,403 training-query records,
for 16,843,945 representations in total. The 128-dimensional representations
encode query text, item titles, prices, keywords, and OCR-derived text with a
distilled BGE encoder aligned by query--query, item--item, and query--item
objectives.

The three- and five-level settings use 1024-512-128 and
1024-512-128-64-64 SIDs, respectively. PRQ-KMeans uses $k=5$ and
$\beta=15$. All tokenizers share BART-Base \cite{lewis2020bart} and the same
three training stages: semantic content alignment with 35,689,588 examples,
co-occurrence synchronization with 54,459,755 examples, and user
personalization modeling with 45,916,427 examples. Training uses batch size
512 and learning rate $5\times10^{-5}$; the three stages run for 4, 3, and 6
epochs, respectively.

Following OneSearch, we construct separate Click and Order test sets from
search logs, each containing 30,000 query--item pairs with the corresponding
observed behavior. Beam search returns 128 SIDs. For each SID, mapped items
are ranked by availability and a composite score derived from historical
click-through rate, conversion rate, click count, and order count, and the
top five items are retained. The resulting item lists are expanded in beam
order, stably deduplicated, and truncated to 50 items. We report HitRate and
MRR under the cutoff defined in Section~\ref{sec:experimental-setup}.

\subsection{Baseline Configurations}

The public and three-level industrial comparisons include RQ-VAE, R3-VAE,
RQ-KMeans, RQ-GMM, PQ-KMeans, and OPQ-KMeans. In the five-level industrial
comparison, RQ-OPQ uses two RQ-KMeans levels, a balanced K-Means third level,
and two OPQ suffix tokens. ResKmeansFSQ uses three RQ-KMeans levels and maps
its L3 residual with a joint 12-bit FSQ code split into two 64-way suffix
tokens. PRQ-OPQ replaces the first two RQ-KMeans levels of RQ-OPQ with
PRQ-KMeans while retaining its balanced third-level K-Means and two OPQ
suffix tokens.

For a fair comparison, each tokenizer uses its prescribed fitting or training
schedule.
RQ-VAE and R3-VAE are trained for 20 epochs with batch size 4,096 and
learning rate $5\times10^{-4}$. RQ-KMeans and PQ-KMeans use 25 clustering
iterations, OPQ-KMeans uses 25 alternating rotation and clustering iterations,
and RQ-GMM uses at most 30 EM iterations with variance floor $10^{-6}$.
PRQ-KMeans uses 25 soft centroid-refinement iterations.

\subsection{Metrics and Reproducibility}
\label{app:implementation}

Let $\mathcal{S}$ be the set of distinct full SIDs assigned to items and
let $n(s)$ be the number of items assigned to $s\in\mathcal{S}$. The
independent-code ratio is
\begin{equation}
\operatorname{ICR}
=
\frac{|\{s\in\mathcal{S}:n(s)=1\}|}{|\mathcal{S}|}.
\end{equation}
Its denominator is the number of distinct SIDs rather than the number of
items. For level $\ell$, let $\mathcal{P}_\ell$ be the set of occupied
length-$\ell$ SID prefixes. Cumulative prefix utilization is
\begin{equation}
\operatorname{Util}_{\ell}
=
\frac{|\mathcal{P}_\ell|}
{\prod_{t=1}^{\ell}K_t}.
\end{equation}
Used-prefix Gini is computed from the item frequencies of prefixes in
$\mathcal{P}_\ell$ only. If
$M_\ell=|\mathcal{P}_\ell|$ and
$f_{(1)}\le\cdots\le f_{(M_\ell)}$ are their sorted item frequencies, then
\begin{equation}
\operatorname{Gini}_\ell
=
\frac{2\sum_{q=1}^{M_\ell}qf_{(q)}}
{M_\ell\sum_{q=1}^{M_\ell}f_{(q)}}
-\frac{M_\ell+1}{M_\ell}.
\end{equation}
Unused prefixes are excluded, and a lower value indicates a more even
distribution among the occupied prefixes. Table~\ref{tab:industry-main}
reports ICR and utilization as percentages. In the three-level block of
Table~\ref{tab:industry-main}, ICR uses the complete three-token SID. In the
five-level block, ICR uses the complete five-token SID, while utilization
and Gini report each tokenizer's L1--L3 prefix.

SID tokenizer fitting, SID encoding, and codebook analyses are conducted on
a CPU server with two AMD EPYC 9654 processors (96 cores per socket) and
2.2 TB RAM. Downstream generative-retrieval training and evaluation are
conducted on a separate server with eight NVIDIA H800 GPUs (80 GB HBM3 each)
and two Intel Xeon Platinum 8558 processors (48 cores per socket).

\section{Additional Experimental Results}
\label{app:additional-experiments}

This section follows the order of the main analysis. It first provides the
carryover measurement and controlled-update details used in the motivation,
then reports the complete public ablations, public-benchmark sensitivity
results, and two qualitative case studies.

\subsection{Residual Carryover Analysis}
\label{app:residual-geometry}

For the industrial RQ-KMeans tokenizer, we average the carryover ratio in
Eq.~\eqref{eq:kuaishou-parallel-fraction} over the full tokenizer-fitting
population. After L2 normalization, the carryover ratio equals the absolute
cosine between the next-level representation and the selected centroid.
For two independent isotropic unit directions, the expected absolute cosine is
\begin{equation}
\mathbb E\!\left[|\mathbf u^\top\mathbf v|\right]
=
\frac{\Gamma(d/2)}
{\sqrt{\pi}\Gamma((d+1)/2)},
\qquad
\mathbf u,\mathbf v
\overset{\mathrm{i.i.d.}}{\sim}
\operatorname{Unif}(\mathbb S^{d-1}).
\label{eq:isotropic-absolute-cosine}
\end{equation}
For $d=128$, this expectation is 7.07\%, providing a dimensional scale for the values in Figure~\ref{fig:carryover-retention-utilization}(a).
Figure~\ref{fig:carryover-retention-utilization}(a) reports
$\bar m^{(\ell)}$ for the three levels.

The controlled update on the industrial dataset used in the main text can be written
directly in terms of centroid subtraction and projection:
let $\vc_i=\vc_{z_i^{(1)}}^{(1)}$, $\rho_i=\norm{\vc_i}$,
$\hat{\vc}_i=\vc_i/\rho_i$,
$\alpha_i=\vr_i^{(1)\top}\hat{\vc}_i$, and
$\vu_i=\vr_i^{(1)}-\alpha_i\hat{\vc}_i$.
\begin{equation}
\begin{aligned}
\vr_i^{(2)}(\eta)
&=
\normalize{
\vr_i^{(1)}-
\left[
\eta\vc_i
+
(1-\eta)
\frac{\vr_i^{(1)\top}\vc_i}
{\norm{\vc_i}^{2}}\vc_i
\right]
}\\
&=
\normalize{
\eta(\alpha_i-\rho_i)\hat{\vc}_i+\vu_i
}.
\end{aligned}
\label{eq:appendix-carryover-retention}
\end{equation}
The second equality follows from Eq.~\eqref{eq:centroid-decomp} and
shows that the operational update scales only the original carryover
term before applying normalization. The L1 codebook and assignments remain
fixed, and L2 is refitted for every value of $\eta$.

% SOURCE [PRELIMINARY_GEOMETRY_TABLES]:
% The Kuaishou mean absolute carryover values are author-provided aggregates
% registered in results/kuaishou/expierment.md pending per-representation
% provenance.
% The carryover-retention control results are author-provided aggregates
% registered under KS-CARRYOVER-RETENTION-UTIL.

\subsection{Public Benchmark Ablations}

Table~\ref{tab:ablation-public} reports the complete public results for the
three ablations, consistent with the main-text ablation findings.

\begin{table}[!ht]
\centering
\scriptsize
\setlength{\tabcolsep}{0.4pt}
\renewcommand{\arraystretch}{1.08}
\begin{tabular*}{\columnwidth}{@{\extracolsep{\fill}}l*{8}{r}@{}}
\toprule
& \multicolumn{2}{c}{Sports} & \multicolumn{2}{c}{Toys} & \multicolumn{2}{c}{Clothing} & \multicolumn{2}{c}{LastFM}\\
\cmidrule(lr){2-3}\cmidrule(lr){4-5}\cmidrule(lr){6-7}\cmidrule(l){8-9}
Variant & Recall & NDCG & Recall & NDCG & Recall & NDCG & Recall & NDCG\\
\midrule
PRQ-KMeans
& \textbf{0.0531} & \textbf{0.0219}
& \textbf{0.0851} & \textbf{0.0361}
& \textbf{0.0382} & \textbf{0.0151}
& \textbf{0.0179} & \textbf{0.0076}\\
$-$ Global
& 0.0505 & 0.0202
& \underline{0.0830} & \underline{0.0359}
& 0.0370 & \textbf{0.0151}
& \underline{0.0107} & \underline{0.0044}\\
$-$ Soft
& \underline{0.0515} & \underline{0.0210}
& 0.0815 & 0.0350
& \underline{0.0371} & \underline{0.0146}
& 0.0043 & 0.0017\\
$-$ Projection
& 0.0503 & 0.0201
& 0.0810 & 0.0345
& 0.0360 & 0.0141
& 0.0099 & \underline{0.0044}\\
\bottomrule
\end{tabular*}
\normalfont\normalsize
\caption{Ablation study on the public benchmarks.}
\label{tab:ablation-public}
\end{table}

\subsection{Public-Benchmark Hyperparameter Sensitivity}
\label{app:sensitivity}

\begin{figure}[H]
  \centering
  \includegraphics[width=\columnwidth]{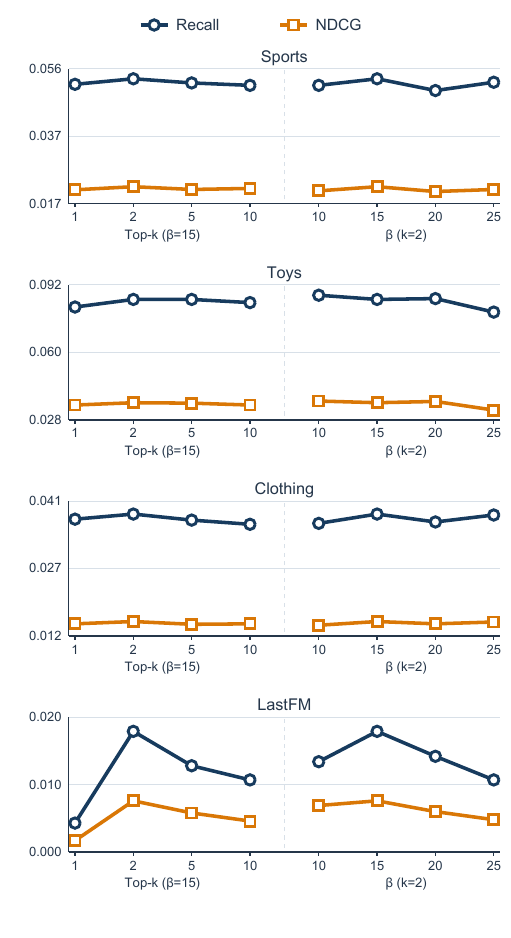}
  \caption{Hyperparameter sensitivity on the public benchmarks.
  Each dataset occupies one panel. The left segment varies Top-$k$ with
  $\beta=15$, and the right segment varies weight concentration with $k=2$.
  Recall and NDCG follow the definitions in
  Section~\ref{sec:experimental-setup}.}
  \label{fig:public-hyperparameter-sensitivity}
\end{figure}

We also evaluate hyperparameter sensitivity on Sports, Toys, Clothing, and
LastFM. Figure~\ref{fig:public-hyperparameter-sensitivity} varies
$k\in\{1,2,5,10\}$ at fixed $\beta=15$ and
$\beta\in\{10,15,20,25\}$ at fixed $k=2$.
For the neighborhood size, $k=2$ gives the best or tied-best Recall and NDCG
on all four benchmarks, while increasing $k$ further provides no consistent
gain. For the concentration parameter, $\beta=15$ performs best on Sports,
Clothing, and LastFM, whereas $\beta=10$ performs best on Toys. Thus, the
preferred concentration varies across datasets.

% SOURCE [PUBLIC_SENSITIVITY]: explicit public sweep runs in
% results/public/parameter_sensitive.md and
% results/public/sensitivity_k2_beta.md. The k=1 point uses the -Soft row in
% tab:ablation-public, and the shared k=2,beta=15 endpoint is anchored to
% tab:public-main. The plotting script excludes the separate default rows,
% Beauty, ML-1M, k=15, and the beta=5 internal endpoint.

\subsection{Qualitative Case Studies}
\label{app:qualitative-prefix}

Tables~\ref{tab:full-sid-collision-case} and~\ref{tab:weak-intent-case}
examine two aspects of hierarchical SID organization. The collision case
follows one fixed item group through the two tokenizers, and the weak-intent
case compares the prefix neighborhoods associated with the same query.

For the collision case, we select a 12-item Kitty-themed slippers group using
a fixed SID-structure criterion. The group shares one RQ-KMeans full SID.
PRQ-KMeans retains one common level-1 prefix,
separates the items into two level-2 prefixes, and produces three full SIDs
with four items each. Table~\ref{tab:full-sid-collision-case} summarizes the
keywords associated with the resulting PRQ-KMeans branches.

\begin{table}[!ht]
\centering
\small
\setlength{\tabcolsep}{2.5pt}
\begin{tabular}{@{}llcp{0.57\columnwidth}@{}}
\toprule
\shortstack{Full-SID\\branch} & L2 & Items & Keywords\\
\midrule
A & A & 4 & Outdoor; dormitory; shower.\\
B & B & 4 & Indoor; soft-soled.\\
C & A & 4 & Bathroom; travel; home/outdoor.\\
\bottomrule
\end{tabular}
\normalfont\normalsize
\caption{Keyword summaries for the selected 12-item collision group under PRQ-KMeans. Branch labels are anonymized.}
\label{tab:full-sid-collision-case}
\end{table}

For the pre-specified query translated as ``hotel,'' we identify the
level-2 prefix assigned by each tokenizer and inspect the catalog items that
share this prefix. The RQ-KMeans prefix contains 22 items and 14 distinct
level-3 branches, while the PRQ-KMeans prefix contains 100 items and 29
branches. The latter neighborhood includes observed hotel-related contexts
such as entrances, bathrooms, corridors, and commercial installations.

\begin{table}[!ht]
\centering
\small
\setlength{\tabcolsep}{7.0pt}
\begin{tabular}{@{}lrr@{}}
\toprule
Method & \shortstack{Items under\\L2 prefix} &
\shortstack{Distinct\\L3 branches}\\
\midrule
RQ-KMeans & 22 & 14\\
PRQ-KMeans & 100 & 29\\
\bottomrule
\end{tabular}
\normalfont\normalsize
\caption{Prefix-neighborhood statistics for the weak-intent query translated as ``hotel.''}
\label{tab:weak-intent-case}
\end{table}
\end{document}